# CCL: Collaborative Curriculum Learning for Sparse-Reward Multi-Agent Reinforcement Learning via Co-evolutionary Task Evolution

Yufei Lin[1][0009-0005-0941-3316], Chengwei Ye[1][0009-0004-2593-3621], Huanzhen Zhang[2][0009-0008-3051-5642], Kangsheng Wang[*3][0009-0009-8392-4148], Linuo Xu[4][0009-0004-2901-6193], Shuyan Liu[5][0009-0004-7641-2623], Zeyu Zhang[6]

[1] Homesite Group Inc.
[2] Chewy Inc.
[3] University of Science and Technology Beijing
[4] Yunnan University of Finance and Economics
[5] Yunnan University
[6] The Australian National University

[*] Corresponding author.

**Abstract.** Sparse reward environments pose significant challenges in reinforcement learning, especially within multi-agent systems (MAS) where feedback is delayed and shared across agents, leading to suboptimal learning. We propose Collaborative Multi-dimensional Course Learning (CCL), a novel curriculum learning framework that addresses this by (1) refining intermediate tasks for individual agents, (2) using a variational evolutionary algorithm to generate informative subtasks, and (3) co-evolving agents with their environment to enhance training stability. Experiments on five cooperative tasks in the MPE and Hide-and-Seek environments show that CCL outperforms existing methods in sparse reward settings.

## 1 INTRODUCTION

Deep Reinforcement Learning (DRL) has shown substantial success in Multi-Agent Systems (MAS), with notable applications in robotics [1, 2], gaming [3], and autonomous driving [4]. Despite this progress, sparse reward environments continue to hinder learning efficiency, as agents often receive feedback only after completing complex tasks. This delayed reward signal limits exploration and makes policy optimization difficult.

To improve exploration under sparse rewards, several strategies have been proposed, including reward shaping [5, 6], imitation learning [7], policy transfer [8], and curriculum learning [9, 10]. These methods aim to strengthen the reward signal and guide agents toward effective behaviors. While effective in single-agent environments, their performance often degrades in MAS, where multiple interacting agents exacerbate environmental dynamics and expand the joint state-action space [11–13].

In response, we propose **Collaborative Multi-dimensional Course Learning (CCL)**, a co-evolutionary curriculum learning framework tailored for sparse-reward cooperative MAS. CCL introduces three core innovations:
(1) It generates agent-specific intermediate tasks using a variational evolutionary algorithm, enabling balanced strategy development.
(2) It models co-evolution between agents and their environment [14], aligning task complexity with agents' learning progress.
(3) It improves training stability by dynamically adapting task difficulty to match agent skill levels.

Through extensive experiments across five tasks in the MPE and Hide-and-Seek (HnS) environments, CCL consistently outperforms existing baselines, demonstrating enhanced learning efficiency and robustness in sparse-reward multi-agent scenarios.

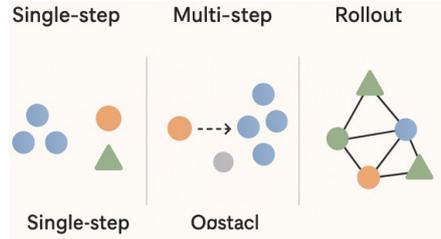

**Fig. 1.** MPE is validated with three different collaborative task scenarios.

## 2    PROBLEM STATEMENT

In reinforcement learning, the reward signal is a critical feedback mechanism guiding agents to assess their actions and learn optimal policies via the Bellman equation [15]. While a well-designed reward function defines the task objective and measures agent behavior, agents may still pursue suboptimal strategies. Nonetheless, carefully crafted rewards greatly enhance learning efficiency and policy convergence [16].

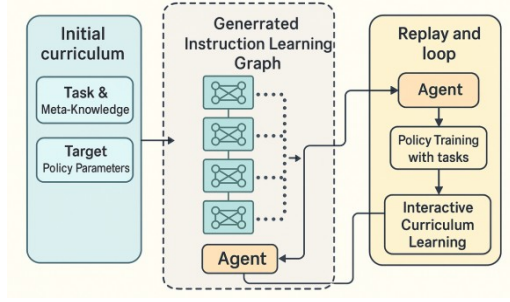

**Fig. 2.** Intermediate task generation in MAS is more complex than in single-agent settings due to the need to account for agent-specific subtasks. In sparse reward environments where rewards are shared, incorporating an individual perspective mechanism becomes essential to ensure effective task decomposition and learning.

Designing dense rewards in complex MAS is challenging due to reliance on prior knowledge, which often fails to capture all interaction dynamics. Sparse rewards offer a more flexible alternative by providing feedback only upon reaching a critical goal state [17], reducing dependence on manual reward design and improving generalization.

In non-sparse reward settings, at each time step $t$, the agent observes its current state $s_t \in S$ and selects an action $a_t \in A$ based on its policy $\pi(a_t|s_t)$. The chosen action results in a transition to a new state $s_{t+1}$, determined by the environment's transition dynamics $p(s_{t+1}|s_t, a_t)$, and an associated reward $r_t$ is obtained from the reward function $r(s_t, a_t, s_{t+1})$. The sequence of states, actions, following states, and rewards over an episode of $T$ time steps form the trajectory $\tau = (s_t, a_t, s_{t+1}, r_t)_{t=0}^{T-1}$, where $T$ is either determined by the maximum episode length or specific task termination conditions. This outlines the process of reinforcement learning for a single agent.

The goal of this individual agent is to learn and maximize its expected cumulative rewarded policy:

$$J = \mathbb{E}_\pi \left[ \sum_t \gamma^t r_t \right] \quad (1)$$

where $\gamma$ is the discount factor, representing future rewards' diminishing value refinement degree of the optimization process is carried out by each time step inside the trajectory, that is, the optimization granularity is accurate to each time step.

However, the system dynamics significantly intensify when extending this general framework to MAS under sparse reward conditions. In this system, there are $N$ decision-making agents, where each agent $i$ takes an action $a_i$ at time step $t$ based on the observed state information and following its dedicated policy $\pi_i(a_i|s_i)$. The global state st of the system is composed of the joint states of all individual agents, denoted as $s_t = (s_1, s_2, \ldots, s_n)$. Correspondingly, the joint action $a_t$ at each time step is also formed by the combination of actions from all agents, i.e., $a_t = (a_1 a_2, \ldots, a_n)$. In the sparse reward environment, reward signals only emerge when the system achieves specific predefined goal states, posing more significant challenges for agent collaboration and strategy optimization.

In cooperative multi-agent tasks, the goal of each agent is no longer focused on maximizing its reward but instead shifts toward optimizing the cumulative reward of the entire system [52]. This requires agents to collaborate effectively, coordinating their actions to achieve the shared objective, thereby improving the overall performance of the multi-agent system. Consequently, the objective function $J$ for each agent $i$ is transformed into $J_i(\pi_i) = \mathbb{E}_{\pi_i}\left[\sum_t \gamma^t r_i(s_t, a_t)\right]$, where $r_i(s_t, a_t)$ represents the reward received by agent $i$ at time step $t$ given the state st and joint action $a_t$. The overall goal of the multi-agent system (MAS) then becomes the sum of the individual objectives, denoted as $J = \sum_i J_i(\pi_i)$.

At this point, it becomes evident that the essence of a multi-agent reinforcement learning algorithm lies in utilizing the rewards earned by all agents to optimize the overall collaborative strategy [53]. However, this challenge is significantly heightened in a sparse reward environment, where agents receive limited feedback, making it difficult to effectively guide their actions and improve coordination toward the collective goal. In the case that there are only very few 0-1 reward signals, the total reward of the system can be simplified to a binary function:

$$r(s_t, g) = \begin{cases} 1, & s_t = g \\ 0, & \text{Otherwise} \end{cases} \quad (2)$$

As the number of agents increases, training variance in MAS grows exponentially. In sparse reward settings, agents must achieve sub-goals aligned with a shared objective, yet often receive little to no feedback, making learning difficult. This lack of guidance hampers exploration and destabilizes training, rendering many single-agent methods ineffective. To address these challenges, we propose Collaborative Multi-dimensional Course Learning (CCL) for more stable and efficient multi-agent training.

$$s_t = g \iff \forall i \in n, s_i = g_i \quad (3)$$

## 3    RELATED WORK

### 3.1    Curriculum Learning

Sparse reward environments have driven the development of various exploration strategies in reinforcement learning, including reward shaping [18], intrinsic motivation [19], and curriculum learning [20]. While the first two enhance learning by densifying rewards, curriculum learning adopts a divide-and-conquer approach—decomposing complex tasks into simpler subtasks arranged in increasing difficulty [1, 21, 22].

In reinforcement learning, curriculum learning involves three main components: task generation, task ranking, and transfer learning [23]. These can be guided by automated methods [9] or expert knowledge, though the latter often introduces biases [24, 25]. Adaptive Automatic Curriculum Learning (ACL) addresses this by dynamically tailoring task sequences to agent progress, without manual intervention [54].

Despite its promise, ACL faces challenges in defining effective evaluation metrics and managing computational cost [26–28]. Current approaches often rely on coarse performance metrics or costly replay mechanisms [29, 30], making it difficult to scale in complex multi-agent settings.

### 3.2 Evolutionary Reinforcement Learning

Evolutionary Algorithms (EAs) optimize policies through selection, mutation, and recombination of candidate solutions based on fitness scores [31]. Their integration with reinforcement learning aims to address issues like sparse rewards and limited policy diversity [29, 30, 34].

Though promising [32, 33], combining EAs with RL introduces challenges, notably the computational overhead from large populations [24] and the difficulty of retaining informative environmental features during evolutionary encoding. Effective integration requires balancing exploration benefits with computational feasibility[44-51].

## 4 METHODOLOGY

### 4.1 The Variational Individual-perspective Evolutionary Operator

In this section, we provide a detailed explanation of all the components of CMCL. As a coevolutionary system with two primary parts, the agents are trained using the existing Multi-Agent Proximal Policy Optimization (MAPPO) algorithm[35], which will not be elaborated on here. The complete workflow of the CMCL algorithm is outlined in Algorithm 1.

Evolutionary Curriculum Initialization Due to the low initial policy performance of a MAS at the start of training, agents struggle to accomplish complex tasks. Therefore, minimizing the norm of task individuals within the initial population is essential. Assuming the initial task domain is $\Omega_0$, the randomly initialized task population should meet the following conditions, where $d$ represents the initial Euclidean norm between the agent and the task, and $\delta$ is a robust hyperparameter, typically set to be about one percent of the total task space size.

$$\frac{1}{|\Omega_0|} \sum_{t_i \in \Omega_0} d(s_i, g_i) < \delta \qquad (4)$$

---

**Algorithm 1** Coevolving Multidirectional Curriculum Learning

**Require:** training episode $N$, curriculum population in episode $i$: $C_i$, total number of tasks in the curriculum population $n_p$, sampling number of tasks $n_t$, initial region of the task $\Omega_0$, soft selection rate $\alpha$, prototype number $k$

1: Initialize $C_0$ by uniform sampling $n_p$ initial tasks in $\Omega_0$
2: Sample $n_t$ tasks from $C_1$
3: Initialize MAS policy $\theta$
4: **for** $i \leftarrow 1$ **to** $N$ **do**
5:    Parallel train MAS on all $n_t$ tasks
6:    $r_j \leftarrow$ success rate on task $j$, $j = 0$ to $n_t$
7:    **Delete bad tasks**
8:    $f_j \leftarrow \frac{1}{1+e^{-2|r_j-0.5|}}$, $j \leftarrow 0$ to $n_t$
9:    $f_{\text{all}} \leftarrow$ **k-prototyped fitness estimator**$(f_1, f_2, \ldots, f_{n_t})$
10:   Initialize empty curriculum population $C_{i+1}$
11:   **for** curriculum pair $c_k, c_{k+\frac{n_p}{2}}$ in $C_i$ **do**
12:      **if** uniform noise $\delta \sim (0,1) > 0.5$ **then**
13:         $kid \rightarrow$ **Multi-directional Cross**$(c_k, c_{k+\frac{n_p}{2}}, f_{\text{all}})$
14:         $C_i \rightarrow C_i + kid$
15:      **else**
16:         $kid \rightarrow$ **Multi-directional Mutate**$(c_k, c_{k+\frac{n_p}{2}}, f_{\text{all}})$
17:         $C_i \rightarrow C_i + kid$
18:      **end if**
19:   **end for**
20:   new task $\rightarrow$ sample $n_t \times \alpha$ tasks on $C_{i+1}$
21:   old task $\rightarrow$ sample $n_t \times (1-\alpha)$ tasks on $C_j$, $j = 0$ to $i$
22:   tasks $\rightarrow$ new task + old task
23:   Update $\theta$ using MAPPO
24: **end for**

---

Task Fitness Definition: Previous methods often assessed intermediate tasks using agent performance metrics [24, 30, 36] or simple binary filters [37], which fail to capture the non-linear nature of task difficulty. Tasks with success rates near 0 or 1 offer little training value, while those closer to the midpoint present a more suitable challenge. To address this, we model fitness as a non-linear function, favoring tasks of moderate difficulty that best support learning progression. To capture this non-linear relationship, we establish a sigmoid-shaped fitness function to describe the adaptability of tasks to the current level of agent performance, where r represents the average success rate of the agents on task $t$.

$$f = \frac{1}{1 + e^{-2|r-0.5|}} \quad (5)$$

Variational Individual-perspective Crossover In a MAS, the single reward signal is distributed across multiple dimensions, especially from the perspective of different agents, leading to imbalances in the progression of individual strategies. Therefore, based on the encoding method mentioned earlier, operating on intermediate tasks at the individual level within the MAS is necessary. Assuming that in a particular round of intermediate task generation, $N$ individuals from the previous task generation $\{T = t_1, t_2, \ldots, t_N\}$ are randomly divided into two groups $T_A$ and $T_B$. Then, we take $N/2$ task pairs $t^A, t^B$ from $T_A$ and $T_B$ to produce new children in the population.

$$\begin{cases} t_i^{A*} \leftarrow t_i^A + S_i \vec{D_i}, \\ t_i^{B*} \leftarrow t_i^B + S_i \vec{D_i}, \end{cases} \quad i = 1, 2, \ldots, \frac{N}{2} \quad (6)$$

In the above formula, $s_i$ represents the crossover step size for pair $i$, and $\vec{D_i}$ represents the crossover direction for pair $i$. The calculations of $s_i$ and $\vec{D_i}$ are shown below:

$$s_{c,i} = \frac{|f(t_i^A) - f(t_i^B)|}{\max(f(T)) - \min(f(T))} \quad (7)$$

$\vec{D_i} = [D_{i,1}, D_{i,2}, \ldots, D_{i,n}]$

$D_{i,j}$ denotes the direction of the $j$-th agent in pair $i$, obtained by uniform random sampling.

$$D_{i,j} = \begin{cases} 0, & \text{if random variable } \delta_j < 0.5 \\ \theta_{i,j}^A - \theta_{i,j}^B, & \text{if random variable } \delta_j \geq 0.5 \end{cases} \quad (8)$$

The proposed variational individual-perspective crossover ensures each agent's subtask direction contributes equally to curriculum evolution, enabling broader exploration compared to traditional methods. In a MAS with nnn entities, this results in 2n2^n2n possible direction combinations, enhancing diversity in intermediate task generation.

To address catastrophic forgetting [38, 39], we adopt a soft selection strategy. Rather than discarding low-fitness individuals, the entire population is retained, and a fraction (α\alphaα, typically 0.2–0.4) of historical individuals is reintroduced each iteration. This maintains task diversity, preserves challenging tasks for future stages, and helps avoid local optima.

### 4.2 Elite Prototype Fitness Evaluation

Evolutionary algorithms often require maintaining a sufficiently large population to ensure diversity and prevent being trapped in local optima or influenced by randomness. However, evaluating the fitness of intermediate tasks in a large curriculum population significantly increases computational cost. To mitigate this issue, we propose a prototype-based fitness estimation method.

First, we uniformly sample tasks in each iteration and measure their success rate r and fitness $f$. These sampled tasks, called prototypes, are used in actual training. Next, we employ a K-Nearest Neighbor (KNN) approach to estimate the fitness of tasks not directly used in training.

Assume there are $m$ individuals in the prototype task set $P$, with fitness values $f_i$ for each $i$, and $n$ individuals in the query task set $Q$, represented as vectors $q_j$ for each $j$. For any individual $q_j$ in the query set $Q$, its fitness value $f_j$ can be calculated as shown below

$$f_j = \frac{1}{k} \sum_{i \in N_j} e_{f_i} \quad (9)$$

In the formula, $N_j$ represents the set of indices of the k-closest individuals in the prototype task set $P$ to the vector $q_j$, based on the Euclidean distance. This can be expressed as follows:

$$N_j = \arg \min_{S \subseteq P, |S|=k} \sum_{i \in S} \|q_j - p_i\|^2 \quad (10)$$

## 5  EXPERIMENT

### 5.1  Main Result

We evaluate CCL on five cooperative tasks across two environments: simple/complex propagation and Push-ball from the MPE benchmark [40], and ramp-passing and lock-back from the MuJoCo-based HnS environment [41]. All tasks use a binary (0-1) sparse reward structure, with results averaged over three random seeds. Training is conducted using MAPPO [35] on a system with an Nvidia RTX 3090 GPU and a 14-core CPU. Attention mechanisms [42] are also integrated to improve agent coordination.

We compare CCL with five baselines:
1. Vanilla MAPPO [35] – Direct training on the target task without intermediate tasks.
2. POET [24] – Uses task evolution; implemented with the same setup as CCL for fairness.
3. GC [36] – An improved version of POET with enhanced task generation.
4. GoalGAN [10] –  Combines curriculum learning with attention-based enhancements.
5. VACL [43] – Applies variational methods to create robust intermediate tasks.

Across all environments, baseline methods struggle under sparse rewards, especially in HnS. CCL consistently outperforms them in both learning speed and final performance, achieving over 95% success in the most complex tasks (see Tables 1 and 2).

### 5.2  Ablation Studies

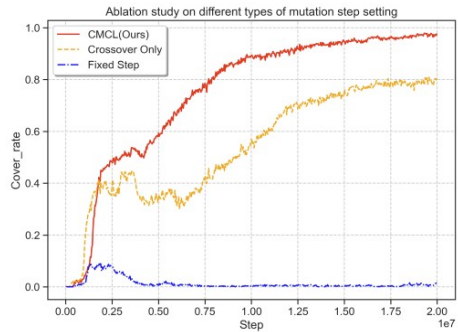

**Fig. 3.** The adaptive step usage ablation experiments which show its effect.

Adaptive Mutation Step: Ablation studies show that using an adaptive mutation step size enhances flexibility and performance in sparse reward environments compared to fixed or no mutation. While mutation promotes strategy diversity, improper step sizes can degrade learning. Notably, adaptive mutation proves as effective as crossover and individual-perspective variation in improving CCL's performance (see Fig. 3).

Non-linear Factor in Fitness Function: As shown in Fig.4, the sigmoid fitness function delivers better performance than the linear form $f = -k|r - 0.5|$. This improvement stems from the sigmoid function's properties: as the agent's success rate approaches 0 or 1, the task's suitability to the agent's abilities decreases exponentially. Specifically, when the success rate is exactly 0.5, the fitness value remains consistently at 0.5. This approach effectively integrates nonlinear elements into the success rate distribution, enabling the fitness function to more accurately represent the relationship between task difficulty and the agent's skill level.

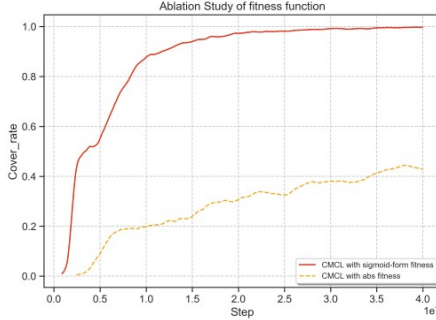

**Fig. 4.** The comparison of using absolute value and sigmoid-shaped fitness function.

**Table 1.** The Performance Comparison of CCL and Other Baselines on Simulated Environments

| Method | Ramp-Use | Lock and Return | Simple-Spread | Hard-Spread | Push-Ball |
|---|---|---|---|---|---|
| | num_agent = 2 | num_box = 2<br>num_agent = 2 | num_agent = 4<br>num_landmark = 4 | num_landmark = 4<br>num_agent = 4 | num_agent = 2<br>num_ball = 2<br>num_landmark = 2 |
| MAPPO[35] | $< 1\%$ | $< 1\%$ | $< 1\%$ | $< 1\%$ | $2\% \pm 0.5\%$ |
| GC[36] | $37.2\% \pm 18.6\%$ | $8.7\% \pm 3.2\%$ | $65\% \pm 12.1\%$ | $79\% \pm 15.6\%$ | $59\% \pm 12.3\%$ |
| POET[24] | $< 1\%$ | $< 1\%$ | $44\% \pm 9.7\%$ | $10\% \pm 8.1\%$ | $80\% \pm 8.4\%$ |
| GoalGAN[10] | $9.2\% \pm 4.2\%$ | $< 1\%$ | $82\% \pm 0.9\%$ | $86\% \pm 8.8\%$ | $61\% \pm 8.7\%$ |
| VACL[43] | $94.7\% \pm 0.8\%$ | $95.4\% \pm 0.1\%$ | $90\% \pm 1.6\%$ | $91\% \pm 6.9\%$ | $90\% \pm 3.0\%$ |
| **CCL (Ours)** | $98.4\% \pm 0.3\%$ | $99.1\% \pm 0.7\%$ | $99\% \pm 0.2\%$ | $95\% \pm 3.4\%$ | $96\% \pm 1.5\%$ |

**Table 2.** Performance Metrics for Various Methods across Different Tasks

| Method | Simple-Spread | Push-Ball | Hard-Spread |
|---|---|---|---|
| MAPPO[35] | $> 5e7$ | $> 1e8$ | $> 1e8$ |
| GC[36] | $> 5e7$ | $> 1e8$ | $1e8$ |
| POET[24] | $> 5e7$ | $> 1e8$ | $1e8$ |
| GoalGAN (att)[10] | $> 5e7$ | $1e8$ | $1e8$ |
| VACL[43] | $> 5e7$ | $1e8$ | $1e8$ |
| **CCL (Ours)** | $2e7$ | $6e7$ | $7e7$ |

## 6    CONCLUSION

This paper presents CCL, a co-evolutionary curriculum learning framework for improving training stability and performance in sparse-reward multi-agent systems (MAS). By generating intermediate tasks, applying variational individual-perspective crossover, and using elite prototype-based fitness evaluation, CCL enhances exploration and coordination. Experiments in MPE and HnS show consistent outperformance over baselines, and ablation studies confirm the value of each component.

While effective in cooperative MAS, future work should extend CCL to competitive or mixed settings. Additionally, storing historical tasks for soft selection increases memory use, which could be mitigated through compression or selective retention.